\documentclass[a4paper,conference]{IEEEtran}
\IEEEoverridecommandlockouts
\usepackage{cite}
\usepackage{amsmath,amssymb,amsfonts}
\usepackage{algorithmic}
\usepackage{graphicx}
\usepackage{textcomp}
\usepackage{xcolor}
\usepackage{bm}
\usepackage{hhline, multirow, colortbl,booktabs}
\usepackage{tikz}
\usepackage{colortbl}
\usepackage[normalem]{ulem}
\usepackage{hyperref}
\DeclareMathOperator*{\argmax}{arg\,max}

\def\BibTeX{{\rm B\kern-.05em{\sc i\kern-.025em b}\kern-.08em
    T\kern-.1667em\lower.7ex\hbox{E}\kern-.125emX}}
    
\newcommand\blfootnote[1]{%
  \begingroup
  \renewcommand\thefootnote{}\footnote{#1}%
  \addtocounter{footnote}{-1}%
  \endgroup
}
    
\begin{document}

\title{PL-$k$NN: A Parameterless Nearest Neighbors Classifier}

\author{\IEEEauthorblockN{Danilo Samuel Jodas}
\IEEEauthorblockA{Department of Computing\\
S\~ao Paulo State University, Brazil\\
danilojodas@gmail.com}
\and
\IEEEauthorblockN{Leandro Aparecido Passos, Ahsan Adeel}
\IEEEauthorblockA{CMI Lab\\School of Engineering and Informatics\\
University of Wolverhampton, UK\\
l.passosjunior@wlv.ac.uk, ahsan.adeel@deepci.org }
\and
\IEEEauthorblockN{Jo\~ao Paulo Papa}
\IEEEauthorblockA{Department of Computing\\
S\~ao Paulo State University, Brazil\\
joao.papa@unesp.br}
}

\maketitle

\begin{abstract}
Demands for minimum parameter setup in machine learning models are desirable to avoid time-consuming optimization processes. The $k$-Nearest Neighbors is one of the most effective and straightforward models employed in numerous problems. Despite its well-known performance, it requires the value of $k$ for specific data distribution, thus demanding expensive computational efforts. This paper proposes a $k$-Nearest Neighbors classifier that bypasses the need to define the value of $k$. The model computes the $k$ value adaptively considering the data distribution of the training set. We compared the proposed model against the standard $k$-Nearest Neighbors classifier and two parameterless versions from the literature. Experiments over 11 public datasets confirm the robustness of the proposed approach, for the obtained results were similar or even better than its counterpart versions.
\end{abstract}


\blfootnote{978-1-6654-9578-3/22/\$31.00 \copyright 2022 IEEE}

\section{Introduction}
\label{s.introduction}

Data classification is the most popular approach in machine learning. The paradigm comprises a set of labeled samples used to train a specific model, i.e., to learn intrinsic characteristics from data, for further classification of unlabeled instances. In this context, one can refer to traditional methods such as the Support Vector Machines~\cite{menezes2019width} and Artificial Neural Networks~\cite{mohammed2020defective}, as well as graph-based models, such as the Optimum-Path Forest~\cite{PapaIJIST09} and the $k$-Nearest Neighbors ($k$-NN)~\cite{Bentley:75}.
$k$-NN is a method used for both classification~\cite{ayyad2019gene} and regression~\cite{luken2019preliminary} purposes. It obtained notorious popularity in the last three decades due to its competitive results in a wide variety of domains, ranging from medicine~\cite{zhong2016predict} to engineering~\cite{farshad2012accurate}, and sentiment analysis~\cite{murugappan2011human}. Although efficient and straightforward, $k$-NN is sensitive to the proper selection of the $k$ value, which may denote a stressful task to the user. A similar problem is faced by most machine learning methods and has been commonly overpassed through metaheuristic optimization algorithms~\cite{passosASOC:19}. Regarding $k$-NN, Wicaksono and Supianto~\cite{wicaksono2018hyper} recently employed such approaches to model the problem of selecting an appropriate $k$. Ling et al.~\cite{ling2015knn} proposed a distance-based strategy to choose the best $k$ by considering a region centered in each instance of the test set. On the other hand, Zhang and Song~\cite{zhang2014knn} proposed a neural network-based model to predict the best $k$ by considering a set of features extracted from each sampled dataset. Besides, Singh et al.~\cite{singh2010pager} proposed a parameterless version of the $k$-NN algorithm for regression purposes. Similar work presented by Desai et al.~\cite{desai2010gear} also claims a parameterless version of $k$-NN for regression purposes. However, the model requires four extra hyperparameters. Ayyad et al.~\cite{ayyad2019gene} proposed a modified version of the $k$-NN classifier for gene expression cancer classification. The proposed model defines a circle with radius $r$ centered at the test sample under prediction, where $r$ is computed by two versions of their Modified $k$-NN (MKNN) model: Smallest MKNN (SMKNN) and Largest MKNN (LMKNN), which measure the minimum and maximum distance between the test sample and each class centroid of the training set, respectively. Although effective in the context of gene data analysis, the method may still suffer from the high amount of neighbors in cases where the class centroids are distant from each other.

This paper proposes the Parameterless $k$-Nearest Neighbors (PL-$k$NN) classifier, a $k$-Nearest Neighbors variant that avoids selecting the proper value of $k$ by introducing a mechanism that automatically chooses the number of neighbors that correctly matches the data distribution. The proposed model is similar to SMKNN presented by Ayyad et al.~\cite{ayyad2019gene}; however, we suggest the following two improvements:

\begin{itemize}
	\item To use the median sample instead of the mean sample as the class centroid;
	\item To use a semicircle whose radius is defined as the distance of the test sample to the nearest class centroid.
\end{itemize}

\noindent Regarding the second contribution, we want to find the training samples assumed to be as close as possible to the nearest centroid of the test sample under prediction. This approach is efficient when there is a mixing of training samples of different classes. In this case, the proposed model defines a semicircle enclosing only the nearest samples close to the cluster assumed as the class of the test instance.

The remainder of the paper is organized as follows: Section II introduces the proposed PL-$k$NN model. Afterward, Sections III and IV present the methodology and the experimental results, respectively. Finally, Section V states conclusions and future work.

\section{Proposed method}
\label{s.propsoed_method}

Let ${\cal Y} = \{\omega_1, \omega_2, \omega_3,\dots,\omega_n\}$ be the set of classes from the dataset, where $\omega_i$ represents the $i^{th}$ class. Also, let ${\cal X} = {\{\bm{x}_{1}, \bm{x}_{2}, \bm{x}_{3},\dots,\bm{x}_{m}\}}$ be the set of samples of the dataset represented by a feature vector denoted as $\bm{x}_{j} \in R^{d}$. In the supervised classification approach, each sample $\bm{x}_j$ is assigned to a class $y_j\in{\cal Y}$ such that the pair $(\bm{x}_{j}, y_{j})$ is used in the subsequent training and testing of the classifier. Formally speaking, this step involves partitioning the samples such that ${\cal X} = {\cal X}_{1}\cup{\cal X}_{2}$ and ${\cal X}_{1}\cap{\cal X}_{2} = \emptyset$, where ${\cal X}_{1}$ and ${\cal X}_{2}$ denote the training and testing sets, respectively.

In the proposed method, the training set is split into $n$ clusters that represent each class $y_i \in {\cal Y}$. Further, the number of nearest neighbors $k$ connected to a target sample $\bm{x}_j$ is adaptively defined according to its distance to all training samples inside the radius of the nearest cluster’s centroid. Let ${\cal C} = \{\bm{c}_1,\bm{c}_2,\ldots,\bm{c}_n\}$ be the set of centroids such that $c_i$ denotes the centroid of the $i^{th}$ cluster, which contains all samples from ${\cal X}_{1}$ that belong to class $\omega_i$. The training stage of the proposed model is summarized as follows:

\begin{enumerate}
	\item Cluster the dataset into $n$ clusters;
	\item For each sample $\bm{x}_{i}\in {\cal X}_{1}$, assign it to the $y_i^{th}$ cluster;
	\item For each $y_i^{th}$ cluster, compute its centroid $\bm{c}_{y_i}$ as the median sample;
	\item Compute the weight of each training sample $\bm{x}_{i}$ regarding its centroid $\bm{c}_{y_i}$ using $W(\bm{x}_{i},\bm{c}_{y_i}) = D_{E}(\bm{x}_{i},\bm{c}_{y_i})^{-1}$, where $D_{E}(\bm{x}_{i},\bm{c}_{y_i})$ denotes the Euclidean distance between sample $\bm{x}_{i}$ and the centroid $\bm{c}_{y_i}$;
	\item Repeat steps 2-4 for all samples in ${\cal X}_1$.
\end{enumerate}

\noindent The training stage is similar to SMKNN, except for the class centroid computation, whose value is computed from the average of each training sample's features in the SMKNN variant.

The PL-$k$NN training stage consists of finding the cluster centroid and the distance weights of all training samples. The distance is computed between each training sample $\bm{x}_{i} \in {\cal X}_{1}$ and the centroid $\bm{c}_{y_i}$ of its cluster. Step 1 is concerned with the splitting of the dataset into $n$ clusters according to the number of classes in ${\cal Y}$. In Step 2, each training sample $\bm{x}_{i}$ is assigned to the cluster of its class $y_{i}$. Step 3 finds the samples assumed to be the center of each cluster by computing its median feature vector. This approach is more effective than using the average due to the following reasons: i) the average is not a valid instance of the cluster, although it is assumed to be located as close as possible to its center, and ii) the average is more sensitive to outliers, and consequently, the resulting value may differ from the data distribution center. In contrast, the median is the sample in the middle of the data distribution, which reduces the effect of instances distant from the dense region of the cluster. Step 4 is defined to assign a weight for all instances inside the cluster. The more distant from the centroid, the smaller the weight of the training sample. Samples distant from the cluster will have less impact as a neighbor of a test sample.

\begin{figure}[!ht]
	\includegraphics[width=1.0\columnwidth]{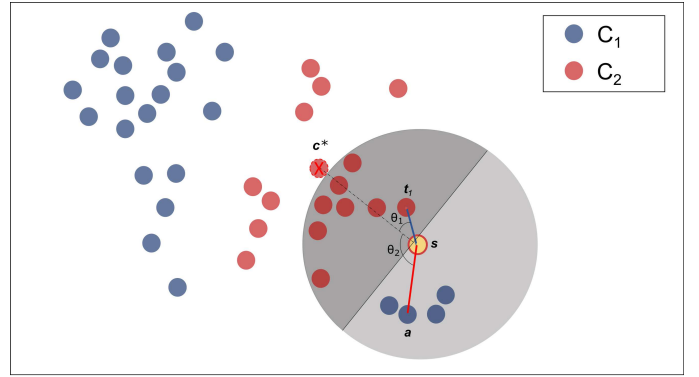}
	\caption{Proposed approach to select the samples inside the semicircle that surrounds the test sample $\bm{s}$. The dimmest area is the semicircle representation proposed for our approach. The dashed line is the distance between $\bm{s}$ and the nearest class centroid $\bm{c}^\star$. The solid blue solid depicts the selected nearest sample $\bm{t}_1$, while the solid red line depicts a sample $\bm{a}$ that is not included in the class prediction of $\bm{s}$.}
	\label{fig.semicircle}
\end{figure}

The following steps are performed for each testing sample $\bm{s}$:

\begin{enumerate}
	\item Calculate the Manhattan distances $D_{M}(\bm{s}, \bm{c}_i)$ between $\bm{s}$ and all cluster's centroids;
	\item Get the centroid $\bm{c}^\star$ with the smallest distance;
	\item Define a circle with radius $D_{M}(\bm{s}, \bm{c}^\star)$ around the test sample $\bm{s}$;
	\item Compute the angle $\theta$ between $\bm{s}$ and $\bm{c}^\star$;
	\item Let ${\cal T}=\{\bm{t}_1,\bm{t}_2,\ldots,\bm{t}_z\}$ be the set of all training samples $\bm{t}_i$ with an angle $\theta_i$ between $-90^o$ and $+90^o$ considering the vector connecting $\bm{s}$ and $\bm{c}^\star$ (dark gray area inside the circle in Figure~\ref{fig.semicircle}). The idea is to pick only the samples inside the semi-circle formed between $\bm{s}$ and the cluster centroid $\bm{c}^\star$;
	\item Determine the final class of $\bm{s}$ as the class with the higher frequency among all training samples $\bm{t}_i\in{\cal T}$ selected in the previous step (see Equations~\ref{eq.prediction} and~\ref{eq.final_class} below).
\end{enumerate}

We want to find the nearest neighbors of the test sample $\bm{s}$ that significantly impact the prediction of its final class. This approach performs similarly to SMKNN, except for the step that chooses the instances that fall inside the circle enclosing $\bm{s}$. Figure~\ref{fig.semicircle} depicts the aforementioned idea.

As illustrated in Figure~\ref{fig.semicircle}, instead of picking all samples inside the circle, we want to find the ones assumed to be as close as possible to the cluster of the centroid $\bm{c}^\star$. Furthermore, since Euclidean distance is sensitive to high-dimensional spaces, we use the Manhattan distance to avoid intensifying features with large differences in such scenarios.


Step 6 of the algorithm above regards the final prediction of the test sample $\bm{s}$ according to the following equations:

\begin{equation}
\label{eq.prediction}
p_i(\bm{s}) = \frac{\displaystyle\sum_{\forall \bm{t}\in{\cal T}\wedge\lambda(t)=\omega_i}{D_{M}(\bm{t}_{j},\bm{s})}^{-1}\ast W(\bm{t}_{j},\bm{c}_i)}{\sum_{k=1}^n p_k(\bm{s})},\forall i\in{\cal Y},
\end{equation}
\begin{center}
and
\end{center}

\begin{equation}
\label{eq.final_class}
y_s = \argmax_i(p_i),\forall i=1,2,\ldots, n,
\end{equation}

\noindent where $y_s$ is the predicted class of $s$, $\lambda(\bm{t})$ outputs the true label of sample $\bm{t}$, and $p_i(\bm{s})$ stands for the probability of sample $\bm{s}$ belonging to class $\omega_i$. We want to penalize the neighbor's samples of $\bm{s}$ with the farthest distance to their cluster centroid. Those samples will have less impact on the final prediction of $\bm{s}$ since they are distant from their correct class group, representing possible outliers.

\section{Methodology}
\label{s.methodology}

This section presents the datasets used in this study and the setup of the experiments.

\subsection{Datasets}

The experiments were performed over 11 public datasets from the UCI Machine Learning~\footnote{\url{https://archive.ics.uci.edu/ml/index.php}} repository. The datasets include binary and multiclass labels, variation in the number of features, and numerical features only. The latter aspect is more concerned with avoiding the encoding of categorical features. The description of the datasets is presented in Table \ref{tab.datasets}.

\begin{table}[!ht]
\centering
\caption{DESCRIPTION OF THE DATASETS USED IN THE EXPERIMENTS.}
\label{tab.datasets}
\begin{tabular}{|l|r|r|r|} 
\hline
\multirow{2}{*}{\begin{tabular}[c]{@{}l@{}}\\\textbf{ Dataset }\end{tabular}} & \multicolumn{3}{c|}{\textbf{ Description }}                                                                                   \\ 
\cline{2-4}
                                                                              & \multicolumn{1}{l|}{\textbf{ Samples }} & \multicolumn{1}{l|}{\textbf{ Features }} & \multicolumn{1}{l|}{\textbf{ Classes }}  \\ 
\hline
Blood Transfusion (BT)                                                        & 748                                     & 4                                        & 2                                        \\ 
\hline
Breast Cancer Diagnostic (BCD)                                                & 569                                     & 30                                       & 2                                        \\ 
\hline
Breast Cancer Original (BCO)                                                  & 699                                     & 10                                       & 2                                        \\ 
\hline
Forest Type (FT)                                                              & 523                                     & 27                                       & 4                                        \\ 
\hline
HCV data (HCV)                                                                & 615                                     & 13                                       & 5                                        \\ 
\hline
Indian Liver (IL)                                                             & 583                                     & 10                                       & 2                                        \\ 
\hline
Mammographic Mass (MM)                                                        & 961                                     & 6                                        & 2                                        \\ 
\hline
Somerville Hapiness (SH)                                                      & 143                                     & 6                                        & 2                                        \\ 
\hline
SPECT Heart (SPTH)                                                            & 267                                     & 44                                       & 2                                        \\ 
\hline
Urban Land Cover (ULC)                                                        & 168                                     & 148                                      & 9                                        \\ 
\hline
Wine (WN)                                                                     & 178                                     & 13                                       & 3                                        \\
\hline
\end{tabular}
\end{table}
\vspace{-0.5cm}
\subsection{Experimental Setup}
The PL-$k$NN model~\footnote{Source code available at https://github.com/danilojodas/PL-kNN.git} and all baselines were implemented using Python 3.6. We rely on Algorithm 1 presented in Ayyad et al.~\cite{ayyad2019gene} to develop the source code of SMKNN and LMKNN. The datasets were divided into 20 folds of training, validation, and testing sets considering a proportion of 70\%, 15\%, and 15\%, respectively. Apart from the statistical analysis between the baselines and the proposed classifier's results, the splitting strategy also enables the optimization of the $k$ value employed by the standard $k$-NN classifier using the training and validation sets. The best $k$ was computed from a range between 1 and 50 to find the value that maximizes the accuracy over the validation set. Also, we used the average accuracy and F1-Score to assess the PL-$k$NN effectiveness. Finally, the Wilcoxon signed-rank~\cite{Wilcoxon45} test with 5\% of significance was employed to evaluate the statistical similarity among PL-$k$NN and the baselines over each dataset. Besides, a post hoc analysis was also conducted using the Nemenyi test~\cite{Nemenyi63} with $\alpha=0.05$ to expose the critical difference (CD) among all techniques.

\section{Experiments}
\label{s.experiments}

This section compares the PL-$k$NN with $k$-NN, SMKNN, and LMKNN. Despite intending the gene classification task, SMKNN and LMKNN are easily adaptable to other contexts due to the cluster analysis, which is intrinsic to any data distribution. Moreover, both techniques work similarly to PL-$k$NN, particularly the SMKNN variant, which is the base of our model, thus constituting a reasonable comparison in public datasets.

For the sake of comparison, the best value for the $k$-NN model was configured using the optimization step described in Section~\ref{s.methodology}. Notice the most accurate average F1-Score is in bold, while similar results according to the Wilcoxon signed-rank test with $5\%$ of significance are underlined for all classifiers. The F1-Score was particularly preferable to assess the model effectiveness because of the imbalanced class distribution of some datasets used in the experiments, such as the Forest Type, HCV data, and SPECT Heart.

Table II presents the average results where the proposed model obtained the best F1-Score in seven out of eleven datasets. The proposed model showed similar and higher F1-Score values for the Blood Transfusion, Forest Type, HCV, Indian Liver, Mammographic, Somerville Happiness, SPECT Heart, Urban Land Cover, and Wine datasets. Even when the proposed model showed inferior results, the average metrics were almost equivalent as observed in the Forest Type and Mammographic Mass datasets. Furthermore, the proposed method surpassed the effectiveness obtained by SMKNN and LMKNN in cases where $k$-NN showed the best F1-Score and the results were statistically different. Notice such behavior presented by Breast Cancer Diagnostic and Breast Cancer Original datasets. Besides, it is worth noting the low performance of LMKNN over some datasets such as Forest Type, HCV, Indian Liver, and Urban Land Cover, which probably relates to its mechanism that assumes a more significant number of neighbors while computing the radius of the circle.

\begin{table}
\centering
\caption{AVERAGE RESULTS OBTAINED BY EACH CLASSIFIER.}
\begin{tabular}{|l|l|l|l|} 
\hline
\multirow{2}{*}{\textbf{ Dataset }} & \multirow{2}{*}{\textbf{Method}} & \multicolumn{2}{c|}{\textbf{ Measure }}                                              \\ 
\cline{3-4}
                                    &                                     & \multicolumn{1}{c|}{\textbf{ Accuracy }} & \multicolumn{1}{c|}{\textbf{ F1-Score }}  \\ 
\hline
\multirow{4}{*}{BT}                 & $k$-NN                                & 0.7844 ± 0.0270                          & 0.3806 ± 0.0934                           \\ 
\cline{2-4}
                                    & LMKNN                               & 0.7594 ± 0.0250                          & 0.3192 ± 0.0786                           \\ 
\cline{2-4}
                                    & SMKNN                               & 0.7585 ± 0.0319                          & \uline{0.4217 ± 0.0944}\uline{}           \\ 
\cline{2-4}
                                    & PL-$k$NN                               & 0.7121 ± 0.0499                          & \textbf{0.4439 }±\textbf{ 0.0848}         \\ 
\hline
\multirow{4}{*}{BCD}                & $k$-NN                                & 0.9553 ± 0.0254                          & \textbf{0.9373 }± \textbf{0.0371}         \\ 
\cline{2-4}
                                    & LMKNN                               & 0.9076 ± 0.0221                          & 0.8595 ±
  0.0376                         \\ 
\cline{2-4}
                                    & SMKNN                               & 0.9247 ± 0.0212                          & 0.8898 ±
  0.0338                         \\ 
\cline{2-4}
                                    & PL-$k$NN                               & 0.9406 ± 0.0176                          & 0.9153 ±
  0.0266                         \\ 
\hline
\multirow{4}{*}{BCO}                & $k$-NN                                & 0.9648 ± 0.0195                          & \textbf{0.9479 }±\textbf{ 0.0300}         \\ 
\cline{2-4}
                                    & LMKNN                               & 0.7743 ± 0.0278                          & 0.5073 ±
  0.0896                         \\ 
\cline{2-4}
                                    & SMKNN                               & 0.9481 ± 0.0196                          & 0.9208 ±
  0.0306                         \\ 
\cline{2-4}
                                    & PL-$k$NN                               & 0.9510 ± 0.0198                          & 0.9255 ±
  0.0310                         \\ 
\hline
\multirow{4}{*}{FT}                 & $k$-NN                                & 0.8692 ± 0.0554                          & \textbf{0.8616 ± 0.0614}                  \\ 
\cline{2-4}
                                    & LMKNN                               & 0.5256 ± 0.0290                          & 0.3014 ±
  0.0219                         \\ 
\cline{2-4}
                                    & SMKNN                               & 0.8513 ± 0.0437                          & 0.8397 ±
  0.0557                         \\ 
\cline{2-4}
                                    & PL-$k$NN                               & 0.8538 ± 0.0424                          & \uline{0.8494 ± 0.0494}\uline{}           \\ 
\hline
\multirow{4}{*}{HCV}                & $k$-NN                                & 0.9136 ± 0.0170                          & \uline{0.4839 ± 0.1036}\uline{}          \\ 
\cline{2-4}
                                    & LMKNN                               & 0.8696 ± 0.0000                          & 0.1860 ±
  0.0000                        \\ 
\cline{2-4}
                                    & SMKNN                               & 0.9071 ± 0.0199                          & \uline{0.4843 ± 0.1245}\uline{}          \\ 
\cline{2-4}
                                    & PL-$k$NN                               & 0.9092 ± 0.0147                          & \textbf{0.4985 ± 0.1197}                  \\ 
\hline
\multirow{4}{*}{IL}                 & $k$-NN                                & 0.6897 ± 0.0378                          & 0.1958
  ± 0.1020                         \\ 
\cline{2-4}
                                    & LMKNN                               & 0.7126 ± 0.0178                          & 0.0723
  ± 0.0643                         \\ 
\cline{2-4}
                                    & SMKNN                               & 0.7121 ± 0.0361                          & 0.2558
  ± 0.0886                         \\ 
\cline{2-4}
                                    & PL-$k$NN                               & 0.7121 ± 0.0294                          & \textbf{0.3336 ± 0.0635}                  \\ 
\hline
\multirow{4}{*}{MM}                 & $k$-NN                                & 0.8045 ± 0.0330                          & \textbf{0.7855 ± 0.0380}                  \\ 
\cline{2-4}
                                    & LMKNN                               & 0.7740 ± 0.0276                          & 0.7627
  ± 0.0276                         \\ 
\cline{2-4}
                                    & SMKNN                               & 0.7823 ± 0.0282                          & 0.7666
  ± 0.0294                         \\ 
\cline{2-4}
                                    & PL-$k$NN                               & 0.7785 ± 0.0302                          & \uline{0.7689 ± 0.0301}\uline{}           \\ 
\hline
\multirow{4}{*}{SH}                 & $k$-NN                                & 0.5476 ± 0.1190                          & \uline{0.4667 ± 0.1844}\uline{}           \\ 
\cline{2-4}
                                    & LMKNN                               & 0.6119 ± 0.0695                          & \uline{0.5106 ± 0.1338}\uline{}           \\ 
\cline{2-4}
                                    & SMKNN                               & 0.5667 ± 0.0877                          & \uline{0.4601 ± 0.1212}\uline{}           \\ 
\cline{2-4}
                                    & PL-$k$NN                               & 0.6167 ± 0.0714                          & \textbf{0.5177 ± 0.1266}                  \\ 
\hline
\multirow{4}{*}{SPTH}               & $k$-NN                                & 0.8087 ± 0.0483                          & 0.3797
  ± 0.1890                         \\ 
\cline{2-4}
                                    & LMKNN                               & 0.7587 ± 0.0582                          & \uline{0.5150 ± 0.1033}\uline{}           \\ 
\cline{2-4}
                                    & SMKNN                               & 0.7163 ± 0.0644                          & 0.4665
  ± 0.0905                         \\ 
\cline{2-4}
                                    & PL-$k$NN                               & 0.7675 ± 0.0507                          & \textbf{0.5320 ± 0.0702}                  \\ 
\hline
\multirow{4}{*}{ULC}                & $k$-NN                                & 0.7782 ± 0.0265                          & 0.7671
  ± 0.0351                         \\ 
\cline{2-4}
                                    & LMKNN                               & 0.5515 ± 0.0390                          & 0.3470
  ± 0.0375                         \\ 
\cline{2-4}
                                    & SMKNN                               & 0.7728 ± 0.0359                          & 0.7654
  ± 0.0393                         \\ 
\cline{2-4}
                                    & PL-$k$NN                               & 0.8005 ± 0.0377                          & \textbf{0.7946 ± 0.0402}                  \\ 
\hline
\multirow{4}{*}{WN}                 & $k$-NN                                & 0.9519 ± 0.0424                          & \uline{0.9530 ±}\uline{ }\uline{0.0424}   \\ 
\cline{2-4}
                                    & LMKNN                               & 0.9481 ± 0.0429                          & \uline{0.9508 ±}\uline{ }\uline{0.0421}   \\ 
\cline{2-4}
                                    & SMKNN                               & 0.9537 ± 0.0368                          & \uline{0.9545 ±}\uline{ }\uline{0.0370}   \\ 
\cline{2-4}
                                    & PL-$k$NN                               & 0.9630 ± 0.0310                          & \textbf{0.9640 }± \textbf{0.0308}         \\
\hline
\end{tabular}
\end{table}

\begin{figure}
	\includegraphics[width=1.0\columnwidth]{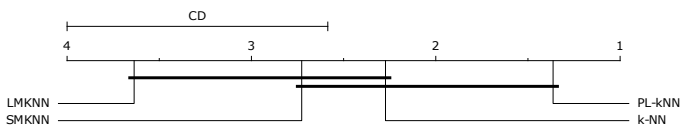}
	\caption{Nemenyi test computed for all techniques.}
	\label{fig.nemenyi_test}
\end{figure}

Besides the Wilcoxon signed-rank test, we also employ the Nemenyi test to provide an overall statistical analysis. The method examines the critical difference among all techniques to plot the method's average rank in a horizontal bar (see Figure~\ref{fig.nemenyi_test}). Notice lower ranks denote better performance, and the methods connected are similar in terms of statistical significance. One can notice the best result attained by the PL-$k$NN model. Also, the proposed approach achieved statistical similarity and better performance compared to the baseline techniques.

\section{Conclusions and Future Works}
\label{s.conclusions}

This paper presented PL-$k$NN, a novel approach for automatically determining the number of nearest neighbors for the $k$-NN classifier. Experiments over 11 datasets showed the competitive results obtained from the proposed model according to statistical analysis applied to all the baselines used for comparison. Besides, the Nemenyi test also confirms the statistical similarity of the PL-$k$NN results with the ones obtained by the standard $k$-NN classifier configured with the best $k$ value.
Regarding future studies, we intend to identify the regions in which the entire circle may be necessary to provide more neighboring samples to increase the prediction accuracy. Furthermore, we also plan to extend PL-$k$NN to regression analysis.

\section*{Acknowledgment}
The authors are grateful to FAPESP grants \#2013/07375-0, \#2014/12236-1, \#2017/02286-0, \#2018/21934-5, \#2019/07665-4, and \#2019/18287-0, Engineering and Physical Sciences Research Council (EPSRC) grant EP/T021063/1, CNPq grants \#307066/2017-7, and \#427968/2018-6, and Petrobras grant \#2017/00285-6.

\bibliographystyle{IEEEtran}
\bibliography{refs}

\end{document}